\documentclass{ecai} 

\usepackage{latexsym}
\usepackage{amssymb}
\usepackage{amsmath}
\usepackage{amsthm}
\usepackage{booktabs}
\usepackage{enumitem}
\usepackage{graphicx}
\usepackage{color}
\usepackage{tcolorbox}
\usepackage{xcolor}
\usepackage{fancyvrb}
\usepackage{adjustbox}
\usepackage{subcaption}
\usepackage{listings}
\usepackage{float}
\usepackage{rotating}
\usepackage{diagbox}
\usepackage{natbib}
\usepackage{amsmath}
\usepackage[utf8]{inputenc}
\usepackage[title]{appendix}
\usepackage{multirow}

\newcommand{\BibTeX}{B\kern-.05em{\sc i\kern-.025em b}\kern-.08em\TeX}

\begin{document}


\begin{frontmatter}


\paperid{752}


\title{Large Language Models Understand Layout}




\author[A]{\fnms{Weiming}~\snm{Li}}
\author[A]{\fnms{Manni}~\snm{Duan}}
\author[A]{\fnms{Dong}~\snm{An}}
\author[A,B]{\fnms{Yan}~\snm{Shao}\thanks{Corresponding Author. Email: shaoyan@cmhi.chinamobile.com.}}

\address[A]{Zhejiang Lab, Hangzhou, China}
\address[B]{China Mobile, Hangzhou Research and Development Center, China}


\begin{abstract}
Large language models (LLMs) demonstrate extraordinary abilities in a wide range of natural language processing (NLP) tasks. In this paper, we show that, beyond text understanding capability, LLMs are capable of processing text layouts that are denoted by spatial markers. They are able to answer questions that require explicit spatial perceiving and reasoning, while a drastic performance drop is observed when the spatial markers from the original data are excluded. We perform a series of experiments with the GPT-3.5/4, Baichuan2, Llama2, and ChatGLM3 models on various types of layout-sensitive datasets for further analysis. The experimental results reveal that the layout understanding ability of LLMs is mainly introduced by the coding data for pre-training, which is further enhanced at the instruction-tuning stage. In addition, layout understanding can be enhanced by integrating low-cost, auto-generated data approached by a novel text game. Finally, we show that layout understanding ability is beneficial for building efficient visual question-answering (VQA) systems.
\end{abstract}

\end{frontmatter}


\section{Introduction}
\vspace{-5pt}
In recent years, large language models (LLMs) have emerged as a dominant force in the global artificial intelligence field, sparking extensive discussions among researchers about their potential and limitations \citep{0,1,2}. Although LLMs are primarily designed for natural language processing (NLP) tasks, some studies demonstrate their additional abilities. For instance, they are employed to generate executable code and even achieve remarkable performance in Google coding interviews \citep{3}.

Beyond text understanding capability, we find that LLMs are capable of processing text layouts that are denoted by spatial markers. As shown in Figure \ref{fig:intro}, we conceptualize the newline-separated plain text as a "visual" two-dimensional canvas, as text editors and browsers are visually two-dimensional intuitively. Three identical questions with distinct answers are arranged in different orientations, interspersed with space markers (denoted as layout). We inquire with ChatGPT about the answers to various orientations. Remarkably, ChatGPT provides accurate responses, and some other open-source LLMs also demonstrate reasonable results. To compare with this, we exclude the space markers from the original data (denoted as strip), resulting in a substantial decline in performance. More examples can be found in Appendix \ref{examples}.

\begin{figure}[htbp]
\rule{\linewidth}{1pt}
\vspace{-20pt}
\begin{center}
    \textbf{(a) Layout}
\end{center}
\vspace{-5pt}
Here are three names mentioned in the context:
\begin{lstlisting}[basicstyle=\footnotesize\ttfamily]
What is your name?    What is your name?
I'm James.            I' m Oliver.

                    What is your name?
                    I'm Emma.
\end{lstlisting}
\textcolor[RGB]{180,0,0}{Question:} What is the name mentioned in the top-left corner? \\
\textcolor[RGB]{0,0,180}{Answer:} The name mentioned in the top-left corner is "James".
\\
\rule{\linewidth}{0.5pt}
\vspace{-20pt}
\begin{center}
    \textbf{(b) Strip}
\end{center}
\vspace{-5pt}
Here are three names mentioned in the context:
\begin{lstlisting}[basicstyle=\footnotesize\ttfamily]
What is your name? What is your name? I'm
James. I' m Oliver. What is your name? I'm
Emma.
\end{lstlisting}
\textcolor[RGB]{180,0,0}{Question:} What is the name mentioned in the top-left corner? \\
\textcolor[RGB]{0,0,180}{Answer:} The name mentioned in the top-left corner is not specified in the given context.
\\
\rule{\linewidth}{1pt}
\caption{Illustration of ChatGPT comprehending text layout.}
\vspace{15pt}
\label{fig:intro}
\end{figure}

This study initiates a comprehensive examination of LLMs' proficiency in understanding text layout, aiming to unravel insights into their performance and implications across various datasets and fine-tuning methodologies.

First, we build up a dataset called TextLayoutQA to evaluate LLMs' text layout understanding capability. Through experiments with the GPT-3.5/4, Baichuan2 \citep{20}, Llama2 \citep{19} and ChatGLM3 \citep{22} models, we uncover that the incorporation of text layout information substantially enhances model performance, resulting in an 8$\sim$33\% gain compared to text without layout.

Furthermore, we explore the effects of pre-training and instruction-tuning stages on LLMs' comprehension of text layout. We illustrate that although LLMs initially demonstrate a basic understanding during pre-training, their proficiency is further enhanced during the instruction-tuning stage.

Moreover, we explore the essential role of training data in shaping LLMs' understanding of text layout, emphasizing the necessity of datasets enriched with layout information, such as code and table data. Through instruction-tuning, we reveal the varying impacts of different types of datasets on LLMs' performance, providing detailed insights into their contributions and constraints.

Our findings not only illuminate the intrinsic capabilities of LLMs in comprehending text layout, but also carry profound implications for their broader applications. By unraveling the intricacies of LLMs' interaction with text layout information, we pave the way for leveraging this capability in tasks ranging from visual question answering (VQA) \citep{31} to document analysis and beyond. Our code and datasets are available on Github.\footnote{https://github.com/liweim/TextLayoutLLM}

The contribution of this paper can be summarized as:
\begin{enumerate}[itemsep=0pt, leftmargin=10pt, parsep=0pt, topsep=0pt, partopsep=0pt]
\item To the best of our knowledge, we are the first to systematically analyze the text layout understanding capability of LLMs.
\item We introduce TextLayoutQA, a dataset designed to assess the text layout understanding capability of LLMs.
\item The origin of text layout understanding capability is thoroughly investigated via instruction-tuning.
\item We propose a low-cost data generation method, approached by a novel text game, that significantly enhances the text layout understanding capability.
\item We show that the text layout understanding capability can be applied in text-rich VQA problems and achieve good performance improvements.
\end{enumerate}


\section{Related Work}
\paragraph{Text layout and language}
Preprocessing for text layout is essential before conducting any NLP on the textual content of intricate documents. \citeauthor{45} \citep{45} present a general approach that integrates language model and text spatial arrangement knowledge. By taking into account text language features and layout characteristics, this method accurately identifies the boundaries and structures of text blocks. However, it relies on rules and is limited to simple cases. Furthermore, it overlooks the relationship between text blocks, which is crucial for document comprehension.

\paragraph{LLMs' capability in text layout understanding}
In recent years, LLMs, represented by the GPT series, have demonstrated strong text comprehension abilities. In this domain, we are aware of some research efforts dedicated to exploring the performance of LLMs in spatial reasoning \citep{0,10,11}, as well as the application of LLMs in graph analysis, understanding, and visualization \citep{1,12,13,14,15}.

A survey by \citeauthor{0} \citep{0} investigates planning and logical reasoning application in spatial environments, finding ChatGPT adept at tracking objects and inferring spatial relationships. Their experiments span various tasks related to physical understanding including optical and shadow projection, spatial viewpoint reasoning, predicting the impact of actions on objects, one-dimensional object sorting, two-dimensional box placement queries, simulated robot exploration like navigating an apartment and searching for a ball in a room, and simulated robot task completion such as setting a table for a meal. However, the effectiveness of these efforts is limited since all "spatial" or "visual" features are interacted in high-level terms. For instance, navigating the apartment involves users informing ChatGPT about the room's location and available doors, followed by describing the new room and door choices after the model makes a selection. In general, ChatGPT demonstrates a certain level of spatial understanding capability, although not in a "geometric" sense.

\citeauthor{1} \citep{1} explores ChatGPT's performance in visual tasks involving ASCII art \citep{18} input, which is an image drawn using ASCII characters in plain text. They find that ChatGPT shows high-level performance in tasks evaluating visual and spatial capabilities, though there is still room for improvement. The study includes tasks like ASCII art recognition, for instance evaluating ChatGPT's ability to handle rotations, scaling, noise, and translation of box plots; ASCII art part identification, such as asking the model about the identity of specific parts of ASCII art images, like heads or tails; and ASCII art generation tasks, for example generating identical copies of ASCII art, removing noise from ASCII art and proportionally enlarging ASCII art.

\paragraph{LLMs' applications in text-rich VQA}
VQA is a task that involves answering questions about images, spanning various formats such as receipts, web pages, tables, documents, or even natural images containing textual information. This task fundamentally requires models to understand multiple modalities. Previous work predominantly focuses on multimodal pre-trained models \citep{32,33,34,35,36}, aiming to leverage all modal information. However, \citeauthor{37} \citep{37} point out that LLMs struggle with text-rich VQA scenarios. The main reasons include the short token lengths, usually 256, of the textual input from the visual encoder to the text encoder in multimodal LLMs, resulting in a significant loss of textual information. Additionally, the low resolution of the image encoder, typically 224*224, compresses and loses a considerable amount of textual information in text-rich images. Under the circumstances, recent work has started exploring the performance of purely textual LLMs in answering questions using only serialized text from images \citep{38}, leveraging the high accuracy of optical character recognition (OCR) models in recognizing long text.


\section{Layout Understanding Based on Spatial Markers}
Text layout refers to the arrangement and presentation of text within a visual space or canvas. It involves the spatial organization of characters, words, and paragraphs to create a visually coherent and aesthetically pleasing display. It is a form of text representation and does not specifically refer to particular data types like codes or tables. Generally, text layout encompasses factors such as newlines, indentation, alignment, font size, and spacing between characters and lines. In this study, we focus on the layout of plain texts, which means the texts do not have formats or font styles. We encode the spatial organization of the texts by using spatial markers such as space and newline, forming a plain text that can be directly input to the LLMs.

In general NLP, text layout is often not explicitly considered because most traditional NLP tasks, such as sentiment analysis, named entity recognition, text classification, and machine translation, focus on understanding the content and meaning of the text rather than its visual presentation. Text layout becomes more relevant when dealing with tasks that involve visual or spatial understanding, such as OCR, document understanding, and certain computer vision tasks. In these cases, the physical placement of text on a page or within an image becomes crucial for accurate interpretation.

However, as research progresses and interdisciplinary approaches become more common, there is an increasing recognition of the importance of text layout understanding, even in traditional NLP. For example, understanding the structure and layout of documents can aid in tasks like information extraction or summarization. As the field evolves, there will be more integration of text layout considerations into a broader range of NLP applications.

The existing datasets that incorporate text layout elements are often related to transcribed documents and tables \cite{30, 31}, but they are not specifically designed to evaluate the text layout understanding capability. Under the circumstances, we introduce a generated dataset called TextLayoutQA specifically for this purpose.

Subsequently, we investigate whether the text layout understanding capability emerges during the pre-training stage or the instruction-tuning stage. Our hypothesis posits that the training corpora with consecutive spaces, including programming code, tables, HTML, and YAML-formatted data, may contribute to the text layout understanding capability of LLMs. To validate this hypothesis, we construct an instruction-tuning dataset that does not include data with consecutive spaces as the instruction-basic dataset. We perform instruction-tuning on LLMs of different types and sizes, utilizing the characteristic of catastrophic forgetting to induce LLMs to "forget" text layout understanding capability. Subsequently, we add training corpora containing consecutive spaces, such as code and table corpora, to the instruction-basic dataset to observe whether the text layout understanding capability is recovered.

Furthermore, inspired by the well-known text game, \textit{word search puzzle}, we devise a novel text game aiming at enhancing text layout comprehension through gameplay. Starting pre-training from scratch with different training corpora would be the most direct validation method. However, due to the substantial computational resources required, it is beyond the scope of our team's capacity and could be considered future work.

Finally, we apply the text layout understanding capabilities of LLMs to text-rich VQA tasks. We introduce a method named textLayoutParser that converts the original texts from VQA datasets to texts with layout. We observe that various LLMs yield better results on text with layout than on those without, highlighting the practical benefits and effectiveness of our research in real-world applications.

\section{Experiments}
\subsection{Datasets}
In this section, we describe all the datasets we use in the experiments. These include three public datasets: XfundQA, FetaQA \citep{30}, DocVQA \citep{31}, along with a generated dataset for text layout understanding evaluation, named TextLayoutQA. Additionally, we propose various instruction-tuning datasets including instruction-basic, instruction-code, instruction-table, instruction-generated, and instruction-test.

\paragraph{XfundQA}
A form QA dataset generated from the XFUND \citep{39} dataset, which is a multilingual form understanding benchmark dataset covering 7 languages (Chinese, Japanese, Spanish, French, Italian, German, Portuguese) with manually annotated forms. Each language includes 199 forms, with 149 forms in the training set and 50 forms in the test set. The dataset involves two sub-tasks: semantic entity recognition and relation extraction. As our primary focus is on QA, we make the following modifications to the Chinese test set of XFUND:

\begin{enumerate}[itemsep=0pt, leftmargin=10pt, parsep=0pt, topsep=5pt, partopsep=0pt]
\item Change the key-value relations to the QA format: "What is the value of the key '\{key\}'?"
\item Remove invalid QA pairs, including those with empty or invalid values and nested key-key-value relations.
\item Rewrite answers with multiple options to the selected one, such as changing "\checkmark A $\square$ B" to "A".
\end{enumerate}

This modified dataset is named XfundQA. Since LLMs' outputs are usually long, we use recall as the evaluation metric, considering a prediction correct if the ground truth appears completely in the LLMs' output.

\paragraph{FetaQA}
A table QA dataset consists of free-form table questions that require deep reasoning and understanding. Most questions are based on discontinuous blocks of information in the table. We conduct evaluations on the test set containing 2,003 samples. Consistently with the dataset's conventions, we use ROUGE-L \cite{rouge} and BLEU-4 \cite{bleu} as the evaluation metrics.

\paragraph{DocVQA}
A document QA dataset consists of printed and typed text as well as scanned documents with various layouts, some of which also include handwritten data. Evaluations are performed on the test set containing 5,188 samples. Following the conventional evaluation, we use average normalized Levenshtein score (ANLS) \citep{31} as the evaluation metric. Since LLMs' outputs are relatively long, the same LLM is used to rephrase the original output answers into shorter ones so that they are aligned with the references.

\paragraph{TextLayoutQA}
A layout QA dataset we proposed to test the layout understanding capability of LLMs. This dataset revolves around enumerating items from various shopping lists arranged in random orientations within the text canvas. As illustrated in Figure 2a, some shopping lists are randomly positioned in four orientations (top-left, top-right, bottom-left, bottom-right) on a newline-separated plain text canvas filled with space markers. Each shopping list is assigned a name (A, B, C, or D) and comprises different products. Both the name and items within the same shopping list are first-letter aligned. For comparison, a version without layout information is constructed for each sample, which involves replacing consecutive space and newline markers with a single space marker. A minimum of two consecutive space markers are maintained between any two shopping lists. Figure 2b illustrates the "without layout" version corresponding to 2a. The paired samples, with and without layout, share the same set of three questions as shown in Figure 2c.

The TextLayoutQA dataset comprises a total of 300 sample pairs, encompassing 900 questions. All questions require the output in list format. F-score is employed to evaluate LLMs' performance. The evaluation process is as follows: first, lists are extracted from the output using regular expressions. Subsequently, the F-score is calculated with each element in the list as a token. If the output does not contain a list, the F-score is calculated with words as tokens, disregarding characters besides words.

\begin{figure}[htbp]
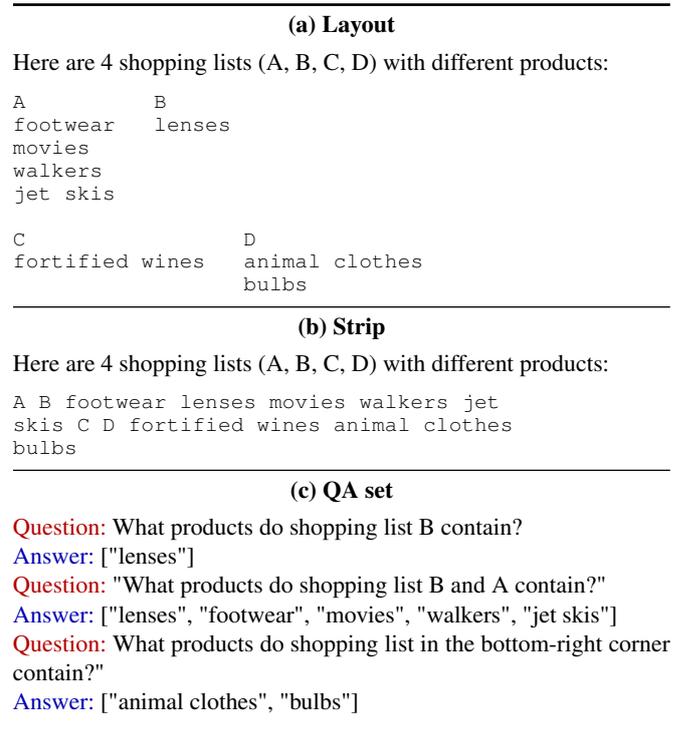

\rule{\linewidth}{1pt}
\vspace{-20pt}
\begin{center}
    \textbf{(a) Layout}
\end{center}
\vspace{-5pt}

Here are 4 shopping lists (A, B, C, D) with different products:\\
\begin{lstlisting}[basicstyle=\footnotesize\ttfamily, aboveskip=-5pt]
A          B
footwear   lenses
movies
walkers
jet skis

C                 D
fortified wines   animal clothes
                  bulbs
\end{lstlisting}

\vspace{-10pt}
\rule{\linewidth}{0.5pt}
\vspace{-20pt}
\begin{center}
    \textbf{(b) Strip}
\end{center}
\vspace{-5pt}
Here are 4 shopping lists (A, B, C, D) with different products:
\begin{lstlisting}[basicstyle=\footnotesize\ttfamily]
A B footwear lenses movies walkers jet
skis C D fortified wines animal clothes
bulbs
\end{lstlisting}

\vspace{-10pt}
\rule{\linewidth}{0.5pt}
\vspace{-20pt}
\begin{center}
    \textbf{(c) QA set}
\end{center}
\vspace{-5pt}
\textcolor[RGB]{180,0,0}{Question:} What products do shopping list B contain? \\
\textcolor[RGB]{0,0,180}{Answer:} ["lenses"] \\
\textcolor[RGB]{180,0,0}{Question:} "What products do shopping list B and A contain?" \\
\textcolor[RGB]{0,0,180}{Answer:} ["lenses", "footwear", "movies", "walkers", "jet skis"] \\
\textcolor[RGB]{180,0,0}{Question:} What products do shopping list in the bottom-right corner contain?" \\
\textcolor[RGB]{0,0,180}{Answer:} ["animal clothes", "bulbs"] \\
\vspace{-5pt}
\rule{\linewidth}{1pt}
\vspace{1pt}
\caption{A pair example of TextLayoutQA dataset with (a) and without layout (b), they share the same QA set (c).}
\label{fig:1}
\vspace{10pt}
\end{figure}

\paragraph{Instruction-basic dataset}
An instruction-tuning dataset designed to diminish the text layout understanding capability of LLMs. Specifically, we randomly select 100k bilingual (English and Chinese) instances from publicly available instruction-tuning datasets \cite{MOSS, Belle, Firefly, Guanaco, hh-rlhf, COIG, hc3, cot, prosocial-dialog, alpacaGPT4}, deliberately excluding consecutive spaces (three or more spaces or two or more tabs), to form the instruction-basic dataset. The distribution of each sub-dataset in the instruction-basic dataset is shown in Table \ref{table:5}.

\begin{table}[htbp]
\caption{Distribution of each sub-dataset in the instruction-basic dataset.}
\centering
\begin{tabular}{c|r r}
\hline
Dataset & Num & Ratio/\% \\
\hline
MOSS & 56,195 & 56.19 \\
belle & 20,881 & 20.88 \\
firefly & 8,929 & 8.92 \\
CSL & 3,289 & 3.28 \\
hh-rlhf & 2,234 & 2.23 \\
COI & 	2,104 & 2.10 \\
HC3 & 1,577 & 1.57 \\
Chain-of-Thought & 1,200 & 1.20 \\
prosocial-dialog & 963 & 0.96 \\
alpacaGPT4 & 851 & 0.85 \\
gpt4tools & 555 & 0.55 \\
GPTeacher & 431 & 0.43 \\
alpaca & 414 & 0.41 \\
webGPT & 173 & 0.17 \\
dolly & 128 & 0.12 \\
Auto-CoT & 59 & 0.05 \\
GAOKAO & 17 & 0.01 \\
\hline
\end{tabular}
\label{table:5}
\end{table}

\begin{table}[htbp]
\caption{Distribution of each sub-dataset in the instruction-code dataset.}
\centering
\begin{tabular}{c|r r}
\hline
Dataset & Num & Ratio/\% \\
\hline
GPT4all & 65,773 & 65.77 \\
CodeAlpaca & 18,911 & 18.91 \\
COIG & 11,048 & 11.04 \\
GPTeacher & 4,268 & 4.26 \\
\hline
\end{tabular}
\label{table:6}
\end{table}

\paragraph{Instruction-code dataset}
An instruction-tuning dataset designed to verify the influence of the code corpora on the text layout understanding capability of LLMs. We randomly sample 100k bilingual (English and Chinese) data from diverse public code-relative instruction-tuning datasets \cite{gpt4all, codealpaca, COIG}. The distribution of each sub-dataset in the instruction-code dataset is shown in Table \ref{table:6}. To preserve the other capabilities of LLMs, these code data are combined with the data from the instruction-basic, resulting in a 200k instruction-code dataset.

\paragraph{Instruction-table dataset}
An instruction-tuning dataset designed to verify the influence of the table corpora on the text layout understanding capability of LLMs. We randomly sample tables from the public table QA dataset, WikiTableQuestions \citep{28}. We introduce text layout by aligning the first characters of all elements in each column of the table using consecutive space markers. A minimum of two consecutive space markers is maintained between any two columns of elements. Distinct from utilizing the dataset's original QA pairs, we reformulate inquiries to elicit the value of each cell in the table, generating 100k new QA pairs. An example is depicted in Figure \ref{fig:2}. These QA instances are combined with the data from the instruction-basic, forming a 200k instruction-table dataset.

\begin{figure}[htbp]
\rule{\linewidth}{1pt}
Given a table: \\
\begin{lstlisting}[basicstyle=\footnotesize\ttfamily, aboveskip=-5pt]
Year       Title         Role
2009-2013  We Speak NYC  Jorge / Fredy
2014-2019  Broad City    Jaime Castro
2015-2016  Alternatino   Arturo
2017       No Activity   Pedro
2019       Alternatino   Arturo
\end{lstlisting}
\textcolor[RGB]{180,0,0}{Question:} What is the Role of Year 2009-2013? \\
\textcolor[RGB]{0,0,180}{Answer:} Jorge / Fredy \\
\rule{\linewidth}{1pt}
\caption{An example of the instruction-table dataset.}
\label{fig:2}
\vspace{10pt}
\end{figure}

\paragraph{Instruction-generated dataset}
An instruction-tuning dataset designed to improve the text layout understanding capability of LLMs. Specifically, we propose a novel text game to generate data automatically, akin to the renowned text game \textit{word search puzzle} (Figure \ref{fig:9}) which challenges to find hidden words within a grid of letters. These puzzles typically feature a rectangular or square grid filled with random letters, accompanied by a list of words to be found. The words can be oriented in various directions—horizontally, vertically, diagonally, and even backward.

\noindent\begin{minipage}{\columnwidth}
\centering
  \includegraphics[width=0.45\columnwidth]{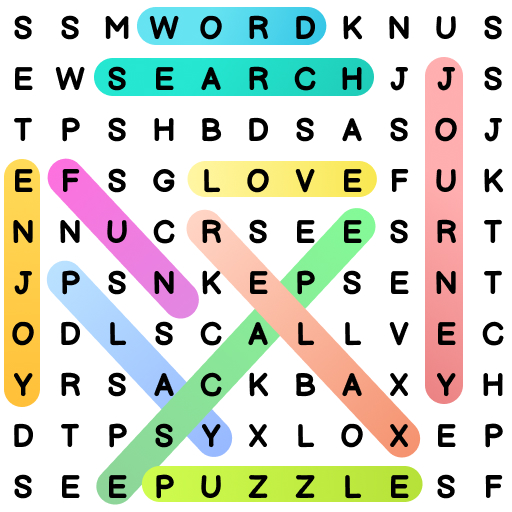}
\captionof{figure}{An example of a word search puzzle.}
\label{fig:9}
\end{minipage}
\vspace{10pt}

\noindent\begin{minipage}{\columnwidth}
\rule{\linewidth}{1pt}
The sentence search puzzle is a game that involves a grid of words, where players are tasked with finding meaningful sentences hidden within the grid.
The challenge lies in locating continuous words that make up meaningful sentences horizontally and vertically.
The unused spaces in the grid are usually filled with random words to add complexity to the puzzle.
Note: answer in the form of a list, for example: ['a', 'b']. If you do not know the answer, reply with the empty list [].
Here is a toy example: \\
\rule{\linewidth}{0.5pt}
\\\\
\includegraphics[width=0.7\columnwidth]{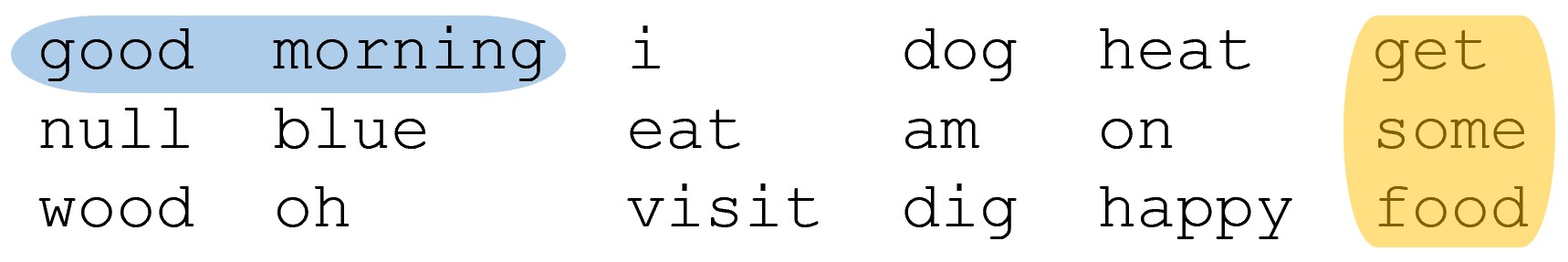}
\rule{\linewidth}{0.5pt}
First, search horizontally and find "good morning". \\
Then, search vertically and find "get some food". \\
So all the sentences hidden in this puzzle are: ["good morning", "get some food"]. \\
Let's solve the following sentence search puzzle step by step: \\
\rule{\linewidth}{0.5pt}
\\\\
\includegraphics[width=\columnwidth]{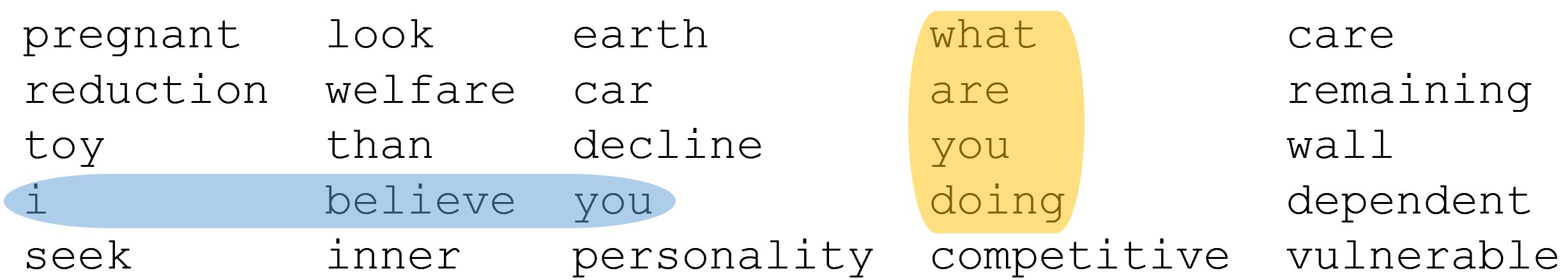}
\rule{\linewidth}{0.5pt}
\\
\textcolor[RGB]{0,0,180}{Answer:}  \\
First, search horizontally and find "i believe you". \\
Then, search vertically and find "what are you doing". \\
So all the sentences hidden in this puzzle are: ['"i believe you"', '"what are you doing"']. \\
\rule{\linewidth}{1pt}
\captionof{figure}{An example of the instruction-generated dataset.}
\label{fig:3}
\end{minipage}
\vspace{10pt}

Acknowledging the scarcity of single letters in training corpora, we adapt the \textit{word search puzzle} into a new game named \textit{sentence search puzzle}. This game is designed to identify hidden sentences within a grid of words, with each word separated by consecutive spaces, and each row maintaining a consistent length. The first letters of all words in each column are aligned. A minimum of two consecutive space markers are maintained between any two columns of elements. The sentences can be oriented in two directions—horizontally and vertically. An illustrative example is provided in Figure \ref{fig:3}.

To mitigate the difficulty of the game, we include intermediate-solving steps in the example provided. We mandate the game to output a list of sentences. During the evaluation step, similar to TextLayoutQA, we employ regular expressions to extract the list from the output.

We randomly generate 100k \textit{sentence search games}. These instances are combined with the data from the instruction-basic, forming a 200k instruction-generated dataset.

\paragraph{Instruction-test}
A testing dataset designed to assess the efficacy of instruction tuning, consisting of four segments of data, as illustrated below, with a total of 800 samples. The evaluation metrics adopted for all instruction-tuning datasets are ROUGE-L and recall.

\begin{itemize}[label=\textbullet, itemsep=0pt, leftmargin=10pt, parsep=0pt, topsep=0pt, partopsep=0pt]
\item Basic: it includes five task types: generation, answering, classification, rewriting, and mathematics, obtained by sampling 100 samples for each task type from the public dataset, natural-instructions \citep{29}, which is a benchmark of 1,616 diverse NLP tasks and expert-written instructions.
\item Code: we randomly select 100 samples from the public LeetCode dataset oa\_leet10k.\footnote{https://huggingface.co/datasets/cognitivecomputations/oa\_leet10k}
\item Table: we construct 100 samples using the same method as building the instruction-table dataset, sourced from the public table QA dataset, FeTaQA \citep{30}.
\item Generate: we generate 100 samples using the same method as building the instruction-generated dataset.
\end{itemize}

\subsection{LLMs}
In our experiments, we select several LLMs for evaluating TextLayoutQA, including GLM3, Llama2, Baichuan2, and GPT-3.5/4. These models cover various sizes and types of open-source and proprietary LLMs that are currently popular.

\begin{itemize}[label=\textbullet]
\item GLM3: the latest generation in the open-source GLM series, featuring a singular parameter size of 6B. The chat version, ChatGLM3-6B exhibits strong performance among pre-training models under 10B, with features such as fluent dialogue and a low deployment threshold.
\item Llama2: the latest open-source LLMs released by Meta, Llama2 surpasses various open-source models on benchmarks and performs comparably to or better than GPT-4 on many test sets. Its open-source license permits commercial utilization, and it comes in a range of parameter sizes — 7B, 13B, and 70B. In our evaluation, we focus on the 7B and 13B parameter sizes.
\item Baichuan2: a new generation open-source LLM introduced by Baichuan Intelligence containing 7B and 13B parameters. Baichuan2 achieves optimal results among models of the same size on various general and domain benchmarks in Chinese, English, and multilingual contexts. Our evaluation encompasses both parameter sizes.
\item GPT-3.5/4: developed by OpenAI, we utilize the GPT-3.5-Turbo and GPT-4-Turbo versions in our experiments. GPT-3.5 is a highly capable model known for generating human-like text based on the input provided. GPT-4 enhances this capability, delivering improved accuracy, comprehension, and contextual understanding.
\end{itemize}

\subsection{Experimental Result and Analysis}
\subsubsection{General Evaluation}
In this section, we first evaluate the text layout understanding capability of LLMs, subsequently analyze how the tokenizers of LLMs encode the consecutive punctuations, and finally compare the performance of using different spatial markers for ablation study.

Table \ref{table:1} shows the evaluation results of different LLMs on TextLayoutQA with (layout) and without (strip) text layout information. Compared to the strip, various LLMs achieve a performance improvement of 8$\sim$33\% with text layout information, indicating the models' ability to understand text alignment, layout, and orientation.

\begin{table}[h]
\caption{Evaluation results, measured by F-score, of different LLMs on TextLayoutQA with (Layout) and without (Strip) text layout.}
\centering
\begin{tabular}{c|cc|c}
\hline
LLMs & Strip & Layout & Difference \\
\hline
ChatGLM3-6B & 33.52 & \textbf{49.52} & +16.00 \\
Llama2-7B & 47.47 & \textbf{58.80} & +11.33 \\
Llama2-13B & 53.45 & \textbf{61.93} & \ \ +8.48 \\
Baichuan2-7B & 47.08 & \textbf{60.82} & +13.74 \\
Baichuan2-13B & 47.00 & \textbf{68.69} & +21.69 \\
GPT-3.5-Turbo & 61.80 & \textbf{87.77} & +25.97 \\
GPT-4-Turbo & 65.42 & \textbf{98.83} & +33.41 \\
\hline
\end{tabular}
\label{table:1}
\end{table}

ChatGLM3, Llama2, Baichuan2, and GPT-3.5/4 all use byte pair encoding (BPE) \citep{23} from SentencePiece \citep{24} as their tokenizer. These tokenizers encode different lengths of consecutive spaces with distinct tokens. Table \ref{table:4} shows that GPT-3.5/4 has distinct tokens for most consecutive punctuation, while ChatGLM3, Llama2, and Baichuan2 only do so for a limited range. However, all these models support encoding relatively long consecutive spaces, suggesting that such spaces are prevalent in their training data. Given that some programming languages like Python and certain file formats like YAML are sensitive to indentation, we assume that these corpora aid LLMs in grasping how consecutive spaces align texts, thereby acquiring text layout understanding capability.

\begin{table}[htbp]
\caption{Maximum lengths of consecutive punctuation that can be encoded by different LLMs' tokenizers.}
\centering
\begin{adjustbox}{width=\columnwidth}
\begin{tabular}{c|c c c c c c}
\hline
LLMs & Space & Tab & Newline & Exclamation & Comma & Full-stop \\
\hline
ChatGLM3 & 15 & 1 & 1 & 2 & 1 & 4 \\
Llama2 & 15 & 1 & 1 & 2 & 1 & 4 \\
Baichuan2 & 32 & 1 & 1 & 1 & 2 & 3 \\
GPT-3.5/4 & 81 & 20 & 14 & 5 & 4 & 9 \\
\hline
\end{tabular}
\end{adjustbox}
\label{table:4}
\end{table}

In TextLayoutQA, we use space and newline as spatial markers. For the ablation study, we investigate another three characters as spatial markers: tab, caron (an accent mark), and a random vanilla character \textit{a}. Notably, newline is still used to separate text lines. Table \ref{table:12} delineates the effects of various LLMs on the TextLayoutQA dataset when deploying these characters as spatial markers. Generally, spaces consistently result in optimal performance for the majority of LLMs. The character \textit{a} generally exhibits the poorest performance across various LLMs due to its lack of spatial semantics. Notably, although caron marker is rare in corpora, it still outperforms strip for most LLMs.


\subsubsection{The Origin of Layout Understanding Capability}
In this section, we first delve into the formation stage of text layout understanding capability and subsequently examine the training corpora used for LLMs. Finally, we explore the type of training corpora that fosters the text layout understanding capability.

\begin{table}[htbp]
\centering
\caption{Evaluation results, measured by F-score, of different markers on encoding text layout information in TextLayoutQA.}
\begin{tabular}{c|c c c c c}
\hline
LLMs & Strip & Space & Tab & Caron & \textit{a} \\
\hline
ChatGLM3-6B & 33.52 & 49.52 & \textbf{49.72} & 37.62 & 16.78 \\
Llama2-7B & 47.47 & \textbf{58.80} & 51.46 & 35.96 & 37.84 \\
Llama2-13B & 53.45 & \textbf{61.93} & 46.00 & 43.39 & 29.33 \\
Baichuan2-7B & 47.08 & \textbf{60.82} & 60.30 & 50.48 & 28.49 \\
Baichuan2-13B & 47.00 & \textbf{68.69} & 66.27 & 56.11 & 44.76 \\
GPT-3.5-Turbo & 61.80 & \textbf{87.77} & 87.36 & 72.86 & 45.79 \\
GPT-4-Turbo & 65.42 & \textbf{98.83} & 98.82 & 93.90 & 85.37 \\
\hline
\end{tabular}
\label{table:12}
\end{table}

Due to the base models' limited ability to strictly follow instructions, their outputs often misalign with QA task references. To ensure a fair comparison of layout understanding between base and chat models, we use perplexity, a common metric for assessing language models, where lower perplexity indicates better performance. Table \ref{table:2} compares the perplexity of different LLMs on the TextLayoutQA dataset with and without layout. All base models show lower perplexity on text with layout, suggesting some inherent text layout understanding from pre-training. After instruction-tuning, chat models demonstrate even lower perplexity on text with layout compared to base models, indicating enhanced layout understanding. To minimize the influence of context length on perplexity, newline markers are used for padding at the beginning of text without layout, with the padding length being the difference between the length of text without layout and the length of text with layout after tokenization.

\begin{table}[h]
\caption{Perplexity of different LLMs on TextLayoutQA dataset with (Layout) and without (Strip) text layout. Lower perplexity indicates better modeling performance.}
\centering
\begin{tabular}{c|c|cc|c}
\hline
LLMs                            & Type & Strip                       & Layout                               & Difference \\ \hline
                                & Base & 6.87                        & \textbf{4.98}                        & -1.89      \\
\multirow{-2}{*}{ChatGLM3-6B}   & Chat & 5.58                        & \textbf{3.56}                        & -2.02      \\ \hline
                                & Base & 2.33                        & \textbf{1.85}                        & -0.48      \\
\multirow{-2}{*}{Llama2-7B}     & Chat & 2.95                        & \textbf{2.26}                        & -0.69      \\ \hline
                                & Base & 2.15 & \textbf{1.81} & -0.34      \\
\multirow{-2}{*}{Llama2-13B}    & Chat & 3.06                        & \textbf{2.27}                        & -0.79      \\ \hline
                                & Base & 1.90                        & \textbf{1.40}                        & -0.50      \\
\multirow{-2}{*}{Baichuan2-7B}  & Chat & 3.09                        & \textbf{1.53}                        & -1.56      \\ \hline
                                & Base & 1.89                        & \textbf{1.33}                        & -0.56      \\
\multirow{-2}{*}{Baichuan2-13B} & Chat & 3.03                        & \textbf{1.35}                        & -1.68      \\ \hline
\end{tabular}
\label{table:2}
\end{table}

Table \ref{table:3} presents the training corpora utilized by various LLMs during the pre-training stage. Notably, the training corpora for GLM3, Llama2, and GPT-4 are not explicitly published, so related information about GLM-130b, Llama, and GPT-3.5 is considered respectively. We find that GLM, Llama, and GPT-3.5 all use datasets such as CommonCrawl, Wikipedia, and Books (Pile includes CommonCrawl, Wikipedia, and Books) in their pre-training. CommonCrawl is a large-scale, unstructured, multilingual web dataset containing over 8 years of web crawler data. Additionally, GLM and Llama utilize code-related sources like GitHub and StackExchange. We do find some examples with various text layouts sourced from GitHub and StackExchange within the Pile dataset. The specific examples can be found in Appendix \ref{corpora}.

\begin{table}[htbp]
\caption{Training corpora used by different LLMs.}
\centering
\begin{adjustbox}{width=\columnwidth}
\begin{tabular}{c|p{7cm}}
\hline
LLMs & Training corpora \\
\hline
GLM & Pile, Chinese WudaoCorpora, Chinese corpora (including online forums, encyclopedia, and QA) crawled from the website \\
\hline
Llama & CommonCrawl \citep{25}, C4 \citep{26}, Github, Wikipedia, Books, ArXiv, StackExchange \\
\hline
Baichuan2 & General internet webpages, books, research papers, codebases, and more \\
\hline
GPT-3.5 & CommonCrawl, WebText \citep{27}, Books1, Books2, Wikipedia \\
\hline
\end{tabular}
\end{adjustbox}
\label{table:3}
\end{table}

We perform instruction-tuning on the instruction-basic, instruction-code, instruction-table, and instruction-generated dataset using Firefly \citep{Firefly} tuning framework. Each dataset is partitioned into training and validation sets with a ratio of 98:2. The training sets undergo 5 epochs with early stopping. As expected, adding different instruction types improves the corresponding ability. The instruction-tuning parameters can be referred to Appendix \ref{parameters}.

Table \ref{table:8} presents the performance of LLMs on TextLayoutQA following various types of instruction-tuning. The comprehensive result can be found in Appendix \ref{instruction-result}. Due to the characteristic of catastrophic forgetting, the text layout capability decreases after instruction-basic tuning compared to the chat model, except for the Llama2 series. This is because the Llama2 chat models tend to produce long responses and sometimes fail to follow the required output format. However, a significant recovery in text layout capability is observed after instruction-code tuning, underscoring the crucial role of code-related data in enhancing text layout understanding. In contrast, the model fine-tuned on instruction-table data experiences a decrease in text layout capability, indicating that the table-related data do not contribute to the text layout capability. It is noteworthy that performance shows enhancement following instruction-generated tuning. For specific small models like Llama2-7B and Baichuan2-7B, instruction-generated tuning even yields the best result among all instruction-tuning datasets. Considering that generating data is significantly more convenient and cost-effective than collecting, we have laid a promising path for pre-training LLMs.

\begin{table}[htbp]
\caption{Evaluation results, measured by F-score, of different LLMs on TextLayoutQA after applying different instruction-tuning datasets. "Origin" indicates no instruction-tuning performed. "Basic", "Code", "Table", and "Generate" correspond to instruction-basic, instruction-table, instruction-code, and instruction-generated datasets, respectively.}
\centering
\begin{tabular}{c|ccccc}
\hline
LLM & Origin & Basic  & Code & Table & Generate \\
\hline
ChatGLM3-6B   & 49.52 & 45.36 & \textbf{63.36} & 28.98 & 59.52 \\
Llama2-7B     & 58.80 & 64.37 & 66.10 & 61.77 & \textbf{66.59} \\
Llama2-13B    & 61.93 & 72.31 & \textbf{73.69} & 66.71 & 66.26 \\
Baichuan2-7B  & 60.82 & 59.93 & 60.77 & 53.79 & \textbf{63.78} \\
Baichuan2-13B & 68.69 & 66.14 & \textbf{72.06} & 65.93 & 69.15 \\
\hline
\end{tabular}
\label{table:8}
\end{table}

\subsubsection{Applications}
In this section, we illustrate the utilization of LLMs' text layout understanding capability in the text-rich VQA domain. We introduce a method named textLayoutParser designed to parse texts with diverse layouts from documents, including plain texts, forms, tables, images, and their combinations. The method involves the placement of text on a two-dimensional character canvas according to the text's coordinates. Detailed implementation is available in Appendix \ref{textLayoutParser}. We evaluate the zero-shot performance on the test sets of three datasets: XfundQA, DocVQA, and FeTaQA. The prompts utilized for each dataset are provided in Appendix \ref{prompt}.

\paragraph{XfundQA}
We use the OCR output provided by the dataset and construct corpora with text layout using textLayoutParser. As a comparison, we replace consecutive spaces and newlines with a single space marker, forming corpora without text layout. The evaluation results of different LLMs on XfundQA with and without text layout are presented in Table \ref{table:9}. Notably, corpora with text layout lead to performance improvements ranging from 0.83\% to 9.55\% compared to corpora without text layout. This infers that the LLM's text layout understanding capability extends to Chinese.

\begin{table*}[htbp]
\centering
\captionof{table}{Evaluation results, measured by ROUGE-L and BLEU-4, of different LLMs on FeTaQA test set with (Layout) and without (Strip) text layout.}
\begin{tabular}{c|ccc|ccc}
\hline
\multirow{2}{*}{LLMs}             & \multicolumn{3}{c|}{ROUGE-L}                               & \multicolumn{3}{c}{BLEU-4}                                \\ \cline{2-7}
                        & {Strip} & {Layout} & Difference & {Strip} & {Layout} & Difference \\ \hline
ChatGLM3-6B                       & {28.79} & {\textbf{31.28}}  & +2.49      & {10.84} & {\textbf{11.08}}  & +0.24      \\ \hline
Llama2-7B                         & {19.71} & {\textbf{24.03}}  & +4.32      & {5.82}  & {\ \ \textbf{7.67}}  & +1.85      \\ \hline
Llama2-13B                         & {27.07} & {\textbf{30.49}}  & +3.42      & {9.14}  & {\textbf{10.91}}  & +1.77      \\ \hline
Baichuan2-7B                     & {32.26} & {\textbf{34.26}}  & +2.00      & {12.15} & {\textbf{13.57}}  & +1.42      \\ \hline
Baichuan2-13B                     & {34.46} & {\textbf{39.15}}  & +4.69      & {14.04} & {\textbf{16.51}}  & +2.47      \\ \hline
GPT-3.5-Turbo                     & {39.05} & {\textbf{39.76}}  & +0.71      & {16.21} & {\textbf{16.63}}  & +0.42      \\ \hline
GPT-4-Turbo                     & {40.29} & {\textbf{41.88}}  & +1.59      & {16.80} & {\textbf{17.77}}  & +0.97      \\ \hline
\end{tabular}
\label{table:10}
\end{table*}

\begin{table*}[htbp]
\centering
\captionof{table}{Evaluation results, measured by ROUGE-L and BLEU-4, of different table encoding methods on FeTaQA test set: "Array," which transforms the original array table data into string format; "Linear," which employs distinct identifiers to differentiate headers and rows; "Triplet," which formats each element as a col-row-value triplet to create a list; and "Ours," which utilizes spaces and newlines to align and separate elements within the table.}
\begin{tabular}{c|cccc|cccc}
\hline
\multirow{2}{*}{LLMs} & \multicolumn{4}{c|}{ROUGE-L}                                                & \multicolumn{4}{c}{BLEU-4}                                                 \\ \cline{2-9}
                      & {Array} & {Linear} & {Triplet} & Ours  & {Array} & {Linear} & {Triplet} & Ours  \\ \hline
ChatGLM3-6B           & {28.79} & {31.01}  & {31.25}   & \textbf{31.28} & {10.84} & {10.85}  & {11.05}   & \textbf{11.08} \\ \hline
Llama2-7B             & {19.71} & {23.69}  & {22.84}   & \textbf{24.03} & {\ \ 5.82}  & {\ \ 7.63}   & {\ \ 6.98}    & \textbf{\ \ 7.67} \\ \hline
Llama2-13B             & {27.07} & {28.80}  & {26.40}   & \textbf{30.49} & {\ \ 9.14}  & {\ \ 9.92}   & {\ \ 9.22}    & \textbf{10.91} \\ \hline
Baichuan2-7B         & {32.26} & {31.87}  & {31.03}   & \textbf{34.26} & {12.15} & {12.21}  & {11.55}   & \textbf{13.57} \\ \hline
Baichuan2-13B         & {34.46} & {\textbf{40.08}}  & {32.57}   & 39.15 & {14.04} & {\textbf{16.94}}  & {12.53}   & 16.51 \\ \hline
GPT-3.5-Turbo         & {39.05} & {35.21}  & {36.88}   & \textbf{39.76} & {16.21} & {14.15}  & {14.96}   & \textbf{16.63} \\ \hline
GPT-4-Turbo         & {40.29} & \textbf{42.44}  & {40.51}   & {41.88} & {16.80} & \textbf{18.23}  & {17.08}   & {17.77} \\ \hline
\end{tabular}
\label{table:13}
\end{table*}

\begin{table}[htbp]
\caption{Evaluation results, measured by recall, of different LLMs on XfundQA with (Layout) and without (Strip) text layout.}
\centering
\begin{tabular}{c|c c|c}
\hline
LLMs & Strip & Layout & Difference \\
\hline
ChatGLM3-6B & 60.13 & \textbf{66.18} & +6.05 \\
Llama2-7B & 57.41 & \textbf{66.96} & +9.55 \\
Llama2-13B & 58.92 & \textbf{66.60} & +7.68 \\
Baichuan2-7B & 64.70 & \textbf{66.66} & +1.96 \\
Baichuan2-13B & 67.38 & \textbf{73.27} & +5.89 \\
GPT-3.5-Turbo & 76.67 & \textbf{77.50} & +0.83 \\
GPT-4-Turbo & 82.54 & \textbf{85.51} & +2.97 \\
\hline
\end{tabular}
\label{table:9}
\end{table}

\paragraph{DocVQA}
We use the OCR output provided by the dataset and construct corpora with text layout using textLayoutParser. For comparison, consecutive spaces and newlines are replaced with a single space marker, forming corpora without text layout. Table \ref{table:11} shows the evaluation results of different LLMs on the DocVQA test set with and without text layout. Compared to corpora without text layout, different LLMs achieved performance improvements of 2.67\% to 4.63\% on corpora with text layout.

\begin{table}[htbp]
\caption{Evaluation results, measured by ANLS, of different LLMs on DocVQA test set with (Layout) and without (Strip) text layout.}
\centering
\begin{tabular}{c|c c|c}
\hline
LLMs & Strip & Layout & Difference \\
\hline
ChatGLM3-6B & 44.60 & \textbf{48.30} & +3.70 \\
Llama2-7B & 38.50 & \textbf{41.81} & +3.31 \\
Llama2-13B & 41.33 & \textbf{44.42} & +3.09 \\
Baichuan2-7B & 33.50 & \textbf{36.17} & +2.67 \\
Baichuan2-13B & 38.75 & \textbf{41.80} & +3.05 \\
GPT-3.5-Turbo & 62.68 & \textbf{66.95} & +4.27 \\
GPT-4-Turbo & 65.00 & \textbf{69.63} & +4.63 \\
\hline
\end{tabular}
\label{table:11}
\end{table}

\paragraph{FeTaQA}
The FeTaQA dataset provides tables in array format, we convert the array table data into string format serving as corpora without text layout. Additionally, corpora refactored by textLayoutParser are used as corpora with text layout. Table \ref{table:10} presents the evaluation results of different LLMs on the FeTaQA test set with and without text layout. Notably, various LLMs showcase performance enhancements ranging from 0.71\% to 4.69\% (ROUGE-L) and 0.24\% to 2.47\% (BLEU-4) on corpora with text layout, compared to those without.

Difference text layout encoding methods are tailored to specific cases. For instance, in the context of table QA, common table encoding techniques include employing identifiers to distinguish headers and rows (referred to as \textit{Linear}) \citep{42,43} and representing each element as a col-row-value triplet to create a list (referred to as \textit{Triplet}) \citep{44}. Apart from our proposed method, we explore several other text layout encoding techniques for an ablation study. Examples of different table encoding methods are available in Appendix \ref{table-encoding}. Table \ref{table:13} provides a performance assessment of various table encoding methods on the FeTaQA test set. Our proposed method outperforms others for ChatGLM3-6B, Llama2-7B, and GPT-3.5-Turbo. Conversely, for Baichuan2-13B and GPT-4-Turbo, the Linear encoding method demonstrates superior results.

\section{Conclusion}
This study extensively investigates the potential of LLMs in text layout understanding by constructing the TextLayoutQA dataset for in-depth research. Experiments utilizing various LLMs demonstrate that, compared to text without layout, the performance of LLMs on datasets with text layout improves by 8$\sim$33\%, confirming their potential in text alignment, layout, and orientation understanding. The additional experiments show that during the pre-training phase, the base models already possess preliminary text layout understanding capabilities, which are further enhanced during instruction tuning. Through ablation experiments with diverse instruction-tuning datasets, we find that training data is crucial for LLMs to acquire text layout understanding, particularly datasets containing text layouts such as codes. In addition, text layout understanding can be enhanced by low-cost auto-generated data approached by a novel text game. Subsequently, leveraging the text layout understanding capabilities of LLMs, we propose an approach named TextLayoutParser to address text-rich VQA problems, achieving decent performance improvements on the XfundQA, FetaQA, and DocVQA datasets.

In summary, our research unveils capabilities in LLMs that have been underexplored, demonstrating their potential to enhance the performance of text-rich VQA problems, expanding the application scenarios of language-centric LLMs, and providing new perspectives for subsequent LLM corpora preparation.


\section{Acknowledgments}
This research was supported by "Pioneer" and "Leading Goose" R\&D Program of Zhejiang (grant no. 2024C01020).



\bibliographystyle{IEEEtran}
\bibliography{reference}

\newpage
\appendix
\setcounter{figure}{0}
\setcounter{table}{0}
\renewcommand{\thefigure}{A\arabic{figure}}
\renewcommand{\thetable}{A\arabic{table}}

\begin{appendices}

\section{Examples of LLMs Text Layout Understanding Capability} \label{examples}
During the early exploration of GPT-3.5-Turbo's QA ability, its remarkable capability to comprehend text alignment, layout, and orientation was discovered. Figure \ref{fig:6} shows some examples of the exploration.

\begin{figure}[htbp]
\rule{\linewidth}{1pt}
\vspace{-20pt}
\begin{center}
    \textbf{Example 1}
\end{center}
\vspace{-15pt}
\begin{verbatim}
                Tom


Jean              Thomas              Lee

                David
\end{verbatim}
\textcolor[RGB]{180,0,0}{Question:} What is the text in the center? \\
\textcolor[RGB]{0,0,180}{Answer:} The text in the center is "Thomas".

\rule{\linewidth}{0.5pt}
\vspace{-20pt}
\begin{center}
    \textbf{Example 2}
\end{center}
\vspace{-5pt}
Here are two bboxes:
\begin{verbatim}
-------
|     |
|     |     ------
|     |     |    |
|     |     |    |
-------     ------
\end{verbatim}
\textcolor[RGB]{180,0,0}{Question:} Which bbox is larger, left or right? \\
Hint: the bbox with more whitespace inside is larger. \\
\textcolor[RGB]{0,0,180}{Answer:} Based on the given information, the left bbox is larger.

\rule{\linewidth}{0.5pt}
\vspace{-20pt}
\begin{center}
    \textbf{Example 3}
\end{center}
\vspace{-5pt}
Now we define the mathematical symbols with their visual representation using a 5x5 matrix, which is made of "0" and "1". For example: \\
The visual representation of 0 is: \\
11111 \\
10001 \\
10001 \\
10001 \\
11111 \\
The visual representation of 1 is: \\
00100 \\
00100 \\
00100 \\
00100 \\
00100 \\
The visual representation of 2 is: \\
11111 \\
00001 \\
11111 \\
10000 \\
11111 \\
\\
\textcolor[RGB]{180,0,0}{Question:} What is the visual representation of 3? \\
\textcolor[RGB]{0,0,180}{Answer:} The visual representation of 3 is: \\
11111 \\
00001 \\
11111 \\
00001 \\
11111 \\
\textcolor[RGB]{180,0,0}{Question:} What is the visual representation of x? \\
\textcolor[RGB]{0,0,180}{Answer:} The visual representation of x is: \\
10001 \\
01010 \\
00100 \\
01010 \\
10001 \\
\rule{\linewidth}{1pt}
\vspace{1pt}
\caption{Showcases of GPT-3.5-Turbo's text layout understanding capability.}
\label{fig:6}
\end{figure}

\section{Corpora with Layout Information on Github and StackExchange} \label{corpora}
By searching for data within Pile that potentially contains text layout information, we discover considerable relevant data from sources like GitHub and StackExchange. Figure \ref{fig:7} shows some examples.

\begin{figure*}[htbp]
\centering
\rule{15cm}{1pt}
\begin{center}
    \textbf{Example 1: from Github}
\end{center}
\vspace{-5pt}
\begin{lstlisting}[basicstyle=\footnotesize\ttfamily]
            <html id=\"top\">
              <head>
                <meta charset=\"utf-8\">
                <title>The Crosswalk Project</title>
                <link rel=\"shortcut icon\" href=\"/assets/favicon.ico\" type=\"image/x-icon\" />
                <link rel=\"icon\" href=\"/assets/favicon.ico\" type=\"image/x-icon\" />
                <script>
                  WebFontConfig = {
                    custom: {
                      families: ['Clear Sans'],
                      urls: ['/css/fonts.css']
                    },
                    google: {
                      families: ['Source Code Pro:n4,n6']
                    },
                    timeout: 2000
                  };
                </script>
              </head>
            </html>
\end{lstlisting}

\rule{15cm}{0.5pt}
\begin{center}
    \textbf{Example 2: from Github}
\end{center}
\vspace{-5pt}
\begin{lstlisting}[basicstyle=\footnotesize\ttfamily]
            /*
             *  Summary:
             *    Selectors for feature type kCursiveConnectionType
             */
            enum {
              kUnconnectedSelector          = 0,
              kPartiallyConnectedSelector   = 1,
              kCursiveSelector              = 2
            };
\end{lstlisting}

\rule{15cm}{0.5pt}
\begin{center}
    \textbf{Example 3: from StackExchange}
\end{center}
\vspace{-5pt}
\centering
\begin{lstlisting}[basicstyle=\footnotesize\ttfamily,basicstyle=\ttfamily\centering]
** LOGGED HOURS **   ** SICK HOURS **        ** RESULT TABLE **
+--------+-------+  +--------+-------+  +--------+-------+-------+
|Name    | Hours |  |Name    | Hours |  |Name    |Hours  |Sick   |
+--------+-------+  +--------+-------+  +--------+-------+-------+
|David   |47     |  |David   |9      |  |David   |47     |9      |
+--------+-------+  +--------+-------+  +--------+-------+-------+
|David   |9      |                      |David   |9      |0      |
+--------+-------+                      +--------+-------+-------+
\end{lstlisting}

\rule{15cm}{0.5pt}
\begin{center}
    \textbf{Example 4: from StackExchange}
\end{center}
\vspace{-5pt}
\centering
\begin{lstlisting}[basicstyle=\footnotesize\ttfamily]
                            Switch flooding when bonding interfaces in Linux
                                              +----+-----+
                                              | Switch 1 | (layer2/3)
                                              +----+-----+
                                                   |
                                              +----+-----+
                                              | Switch 2 |
                                              +----+-----+
                                                   |
                                        +----------+----------+
              +-------------------------+       Switch 3      +-------------------------+
              |                         +----+-----------+----+                         |
              |                              |           |                              |
              |                              |           |                              |
              |     eth0 (B0:B0:B0:B0:B0:B0) |           | eth4 (B4:B4:B4:B4:B4:B4)     |
              |                         +----+-----------+----+                         |
              |                         |        Host B       |                         |
              |                         +----+-----------+----+                         |
              |     eth1 (B1:B1:B1:B1:B1:B1) |           | eth5 (B5:B5:B5:B5:B5:B5)     |
              |                              |           |                              |
              |                              |           |                              |
              +------------------------------+           +------------------------------+
\end{lstlisting}

\rule{15cm}{1pt}
\vspace{10pt}
\caption{An example of data with layout information on Github and StackExchange.}
\label{fig:7}
\end{figure*}

\section{Parameters of the instruction-tuning} \label{parameters}
\begin{table}[!htbp]
\centering
\caption{Key parameters of the instruction-tuning}
\begin{tabular}{l|l}
\hline
parameter      & value         \\ \hline
num\_train\_epochs      & 5                      \\
learning\_rate          & 1e-4                   \\
max\_seq\_length        & 1024                   \\
lr\_scheduler\_type     & constant\_with\_warmup \\
warmup\_steps           & 1000                   \\
lora\_rank              & 64                     \\
lora\_alpha             & 16                     \\
lora\_dropout           & 0.05                   \\
gradient\_checkpointing & true                   \\
optim                   & adamw\_torch           \\
fp16                    & true                   \\
weight\_decay           & 0                      \\
max\_grad\_norm         & 0.3                    \\ \hline
\end{tabular}
\label{table:14}
\end{table}

\begin{table*}[htbp]
\caption{Test results, measured by ROUGE-L and recall, of using different types of instructions to tune LLMs. The "Instructions" column specifies the different instruction-tuning datasets. "Origin" indicates no instruction-tuning performed. "Base", "Code", "Table", and "Generate" correspond to instruction-base, instruction-table, instruction-code, and instruction-generated datasets, respectively. The "Tasks" row specifies the different subsets within the test set, with "Others" encompasses results excluding "Code", "Table", and "Generate" subsets.}
\centering
\begin{tabular}{c|c|rrrr|rrrr}
\hline
\diagbox{LLMs}{Metrics} &            & \multicolumn{4}{c|}{ROUGE-L} & \multicolumn{4}{c}{Recall} \\ \hline
                            & \diagbox{Instructions}{Tasks}   & \multicolumn{1}{c|}{Code}  & \multicolumn{1}{c|}{Table} & \multicolumn{1}{c|}{Generate} & Others & \multicolumn{1}{c|}{Code}  & \multicolumn{1}{c|}{Table} & \multicolumn{1}{c|}{Generate} & Others \\ \hline
\multirow{5}{*}{ChatGLM3-6B}   & Origin       & \multicolumn{1}{r|}{26.65}          & \multicolumn{1}{r|}{20.53}          & \multicolumn{1}{r|}{8.14}           & 28.49                       & \multicolumn{1}{r|}{67.78}          & \multicolumn{1}{r|}{30.96}          & \multicolumn{1}{r|}{24.51}          & 38.12                       \\
                               & Base         & \multicolumn{1}{r|}{9.25}           & \multicolumn{1}{r|}{25.65}          & \multicolumn{1}{r|}{5.39}           & 33.88                       & \multicolumn{1}{r|}{19.39}          & \multicolumn{1}{r|}{29.25}          & \multicolumn{1}{r|}{11.04}          & 37.15                       \\
                               & Code         & \multicolumn{1}{r|}{20.67} & \multicolumn{1}{r|}{17.38}          & \multicolumn{1}{r|}{7.46}           & 30.71                       & \multicolumn{1}{r|}{36.24} & \multicolumn{1}{r|}{20.65}          & \multicolumn{1}{r|}{16.84}          & 36.04                       \\
                               & Table        & \multicolumn{1}{r|}{2.13}           & \multicolumn{1}{r|}{32.00} & \multicolumn{1}{r|}{5.87}           & 21.49                       & \multicolumn{1}{r|}{3.72}           & \multicolumn{1}{r|}{32.00} & \multicolumn{1}{r|}{6.05}           & 25.37                       \\
                               & Generate     & \multicolumn{1}{r|}{4.79}           & \multicolumn{1}{r|}{15.05}          & \multicolumn{1}{r|}{64.33} & 33.92                       & \multicolumn{1}{r|}{7.14}           & \multicolumn{1}{r|}{15.98}          & \multicolumn{1}{r|}{67.16} & 38.31                       \\ \hline
\multirow{5}{*}{LlaMA2-7B}     & Origin       & \multicolumn{1}{r|}{19.28}          & \multicolumn{1}{r|}{24.34}          & \multicolumn{1}{r|}{0.07}           & 17.41                       & \multicolumn{1}{r|}{66.92}          & \multicolumn{1}{r|}{40.44}          & \multicolumn{1}{r|}{0.11}           & 38.17                       \\
                               & Base         & \multicolumn{1}{r|}{10.59}          & \multicolumn{1}{r|}{25.63}          & \multicolumn{1}{r|}{9.83}           & 32.25                       & \multicolumn{1}{r|}{26.69}          & \multicolumn{1}{r|}{45.40}          & \multicolumn{1}{r|}{27.04}          & 39.52                       \\
                               & Code         & \multicolumn{1}{r|}{31.23} & \multicolumn{1}{r|}{40.50}          & \multicolumn{1}{r|}{8.06}           & 27.61                       & \multicolumn{1}{r|}{68.58} & \multicolumn{1}{r|}{46.24}          & \multicolumn{1}{r|}{18.31}          & 35.36                       \\
                               & Table        & \multicolumn{1}{r|}{23.32}          & \multicolumn{1}{r|}{62.72} & \multicolumn{1}{r|}{5.73}           & 32.91                       & \multicolumn{1}{r|}{54.87}          & \multicolumn{1}{r|}{63.17} & \multicolumn{1}{r|}{17.62}          & 37.96                       \\
                               & Generate     & \multicolumn{1}{r|}{12.53}          & \multicolumn{1}{r|}{30.74}          & \multicolumn{1}{r|}{74.98} & 29.26                       & \multicolumn{1}{r|}{31.13}          & \multicolumn{1}{r|}{37.94}          & \multicolumn{1}{r|}{85.97} & 35.58                       \\ \hline
\multirow{5}{*}{LlaMA2-13B}    & Origin       & \multicolumn{1}{r|}{21.27}          & \multicolumn{1}{r|}{18.97}          & \multicolumn{1}{r|}{0.39}           & 17.68                       & \multicolumn{1}{r|}{68.79}          & \multicolumn{1}{r|}{47.94}          & \multicolumn{1}{r|}{0.44}           & 38.97                       \\
                               & Base         & \multicolumn{1}{r|}{7.47}           & \multicolumn{1}{r|}{47.07}          & \multicolumn{1}{r|}{7.83}           & 35.61                       & \multicolumn{1}{r|}{15.11}          & \multicolumn{1}{r|}{53.50}          & \multicolumn{1}{r|}{16.22}          & 39.69                       \\
                               & Code         & \multicolumn{1}{r|}{30.27} & \multicolumn{1}{r|}{43.65}          & \multicolumn{1}{r|}{5.18}           & 32.79                       & \multicolumn{1}{r|}{68.09} & \multicolumn{1}{r|}{53.55}          & \multicolumn{1}{r|}{16.26}          & 39.81                       \\
                               & Table        & \multicolumn{1}{r|}{12.41}          & \multicolumn{1}{r|}{61.83} & \multicolumn{1}{r|}{8.29}           & 32.25                       & \multicolumn{1}{r|}{25.87}          & \multicolumn{1}{r|}{62.07} & \multicolumn{1}{r|}{11.84}          & 38.75                       \\
                               & Generate     & \multicolumn{1}{r|}{17.53}          & \multicolumn{1}{r|}{43.23}          & \multicolumn{1}{r|}{77.05} & 34.13                       & \multicolumn{1}{r|}{32.70}          & \multicolumn{1}{r|}{46.67}          & \multicolumn{1}{r|}{80.18} & 39.93                       \\ \hline
\multirow{5}{*}{Baichuan2-7B}  & Origin       & \multicolumn{1}{r|}{27.34}          & \multicolumn{1}{r|}{18.05}          & \multicolumn{1}{r|}{3.79}           & 22.72                       & \multicolumn{1}{r|}{70.00}          & \multicolumn{1}{r|}{47.98}          & \multicolumn{1}{r|}{5.90}           & 41.59                       \\
                               & Base         & \multicolumn{1}{r|}{27.81}          & \multicolumn{1}{r|}{18.12}          & \multicolumn{1}{r|}{2.60}           & 22.34                       & \multicolumn{1}{r|}{70.62}          & \multicolumn{1}{r|}{47.48}          & \multicolumn{1}{r|}{5.53}           & 40.95                       \\
                               & Code         & \multicolumn{1}{r|}{26.83}          & \multicolumn{1}{r|}{17.30}          & \multicolumn{1}{r|}{4.32}           & 22.39                       & \multicolumn{1}{r|}{69.97}          & \multicolumn{1}{r|}{47.98}          & \multicolumn{1}{r|}{9.49}           & 40.74                       \\
                               & Table        & \multicolumn{1}{r|}{27.83}          & \multicolumn{1}{r|}{61.29} & \multicolumn{1}{r|}{10.19}          & 25.31                       & \multicolumn{1}{r|}{55.01}          & \multicolumn{1}{r|}{61.67} & \multicolumn{1}{r|}{12.30}          & 37.64                       \\
                               & Generate     & \multicolumn{1}{r|}{12.60}          & \multicolumn{1}{r|}{11.40}          & \multicolumn{1}{r|}{77.64} & 26.65                       & \multicolumn{1}{r|}{31.69}          & \multicolumn{1}{r|}{36.56}          & \multicolumn{1}{r|}{81.90} & 39.49                       \\ \hline
\multirow{5}{*}{Baichuan2-13B} & Origin       & \multicolumn{1}{r|}{29.80}          & \multicolumn{1}{r|}{13.92}          & \multicolumn{1}{r|}{4.22}           & 26.22                       & \multicolumn{1}{r|}{67.04}          & \multicolumn{1}{r|}{46.87}          & \multicolumn{1}{r|}{10.15}          & 41.38                       \\
                               & Base         & \multicolumn{1}{r|}{8.68}           & \multicolumn{1}{r|}{17.98}          & \multicolumn{1}{r|}{9.89}           & 25.49                       & \multicolumn{1}{r|}{21.40}          & \multicolumn{1}{r|}{43.46}          & \multicolumn{1}{r|}{17.49}          & 36.91                       \\
                               & Code         & \multicolumn{1}{r|}{20.58} & \multicolumn{1}{r|}{17.96}          & \multicolumn{1}{r|}{8.04}           & 27.08                       & \multicolumn{1}{r|}{51.82} & \multicolumn{1}{r|}{47.94}          & \multicolumn{1}{r|}{21.68}          & 36.64                       \\
                               & Table        & \multicolumn{1}{r|}{10.72}          & \multicolumn{1}{r|}{63.76} & \multicolumn{1}{r|}{13.73}          & 29.03                       & \multicolumn{1}{r|}{26.97}          & \multicolumn{1}{r|}{64.24} & \multicolumn{1}{r|}{17.93}          & 41.76                       \\
                               & Generate     & \multicolumn{1}{r|}{15.91}          & \multicolumn{1}{r|}{19.52}          & \multicolumn{1}{r|}{75.77} & 24.08                       & \multicolumn{1}{r|}{37.57}          & \multicolumn{1}{r|}{41.45}          & \multicolumn{1}{r|}{80.28} & 35.48                       \\ \hline
\end{tabular}
\label{table:7}
\end{table*}

\section{Test results of using different types of instructions to tune LLMs} \label{instruction-result}
The test results of using different types of instructions to tune LLMs are presented in Table \ref{table:7}. Given LLMs often produce long responses that don't align with the ground truth of the table subset, recall yields more reasonable results than ROUGE-L. It can be observed that, compared to the chat model, the code capability significantly decreases after tuning on instruction-base, but it substantially recovers after tuning on the instruction-code, except for Baichuan2-7B. In contrast to the model tuned on the instruction-base, table capability gains a considerable improvement after tuning on the instruction-table, while layout capability obtains a remarkable improvement following instruction-generated tuning.

\section{Method of textLayoutParser} \label{textLayoutParser}
The implementation of textLayoutParser includes four steps: text parsing, determination of unit character size and coordinate conversion, filling text into the character matrix, and conversion of character matrix to plain text.

\paragraph{Text Parsing}
Utilize appropriate parsing methods based on different file formats to obtain text content and their corresponding positional coordinates. For example, OCR can be used to extract text and coordinates from images, while the PyMuPDF Python library can be employed to parse PDF files. As for table data, we generated bounding boxes (bboxes) for each element in the table, including header and cell, based on coordinates and text length. The generation process is as follows: Each character was treated as a unit character, with an assumption of a spacing of 2 between adjacent elements in the same row and 1 between adjacent elements in the same column.  The maximum text length for all elements in the $j^{th}$ column is denoted as $l$, and the bbox for the $i^{th}$ row's ${j-1}^{th}$ column element ($V_{ij}$ for short) is represented as $[x_1, i, x_2, i+1]$. Then, the bbox coordinate for $V_{ij}$ is $[x_2+2, i, x_2+2+l, i+1]$.

\paragraph{Determination of Unit Character Size and Coordinate Conversion}
Determine a unit character size by analyzing the sizes of all text characters, filtering out characters smaller than this unit size. The other text coordinates are then converted using this unit character size. Define a text $t$ with length n and bbox coordinates $(x_1, y_1, x_2, y_2)$. The approximate character width and height can be calculated as $(x_2-x_1)/n$ and $y_2-y_1$, respectively. Let the unit character's width be $x_0$, and the height be $y_0$. The coordinates for $t$ after conversion become $(x_1/x_0, y_1/y_0, x_2/x_0, y_2/y_0)$, rounded to the nearest integer.

\paragraph{Filling Text into the Character Matrix}
Using the coordinates, insert the text into a character matrix. Initialize a matrix with spaces as elements, setting the rows and columns to the maximum y-value and x-value after conversion of text coordinates. Then, sequentially place the text into the corresponding indices of the matrix from left to right to ensure text continuity. For example, if the converted text coordinate is $(10,10,20,20)$, and the text length is 5, each character of the text is placed in the matrix indices $(10,10)$ to $(15,10)$ one by one.

\paragraph{Conversion of Character Matrix to Plain Text}
Convert the character matrix into the plain text for LLMs. This process involves joining all characters in each row into one line of text, and then combining all lines of text using a newline character as a separator. In order to reduce the redundancy of the dense spaces and newline markers, we remove the first column of those with at least three consecutive columns entirely filled with spaces, replace entire rows filled with spaces with a newline character, and replace at least three consecutive newline markers with two newline markers.

\section{Prompt Designs for Difference Datasets} \label{prompt}
Figure \ref{fig:8} illustrates the prompt designs for different datasets. (a) display one-shot prompting for TextLayoutQA. (b)$\sim$(d) display zero-shot prompting for XfundQA, DocVQA, and FeTaQA, respectively. (e) illustrates the 3-shot prompting for rephrasing answers in the DocVQA dataset. The instructions remain consistent across all LLMs except for the Llama2 series, as depicted in (f). Regarding LLM parameter settings, we utilize a temperature of 0.1, maximum output length of 512, top p of 0.85, and repetition penalty of 1.05.
\rule{\linewidth}{1pt}
\vspace{-20pt}
\begin{center}
    \textbf{(a) An example of one-shot prompting for TextLayoutQA dataset}
\end{center}
\vspace{-5pt}
Given some shopping lists with different products, you are supposed to enumerate the products of specific lists
and answer questions in the form of a list, for example: ['a', 'b'], reply with the list only! If you don't know the answer, reply with the empty list []. \\
\\
For example: \\
Here are 2 shopping lists (A, B) with different products:
\begin{lstlisting}[basicstyle=\footnotesize\ttfamily]
A       B
apple   fish
banana  chair
car
\end{lstlisting}
\textcolor[RGB]{180,0,0}{Question:} What products do shopping list B contain? \\
\textcolor[RGB]{0,0,180}{Answer:} ['fish', 'chair'] \\
 \\
Now answer the question below: \\
\{context\} \\
 \\
\textcolor[RGB]{180,0,0}{Question:} \{question\} \\
\textcolor[RGB]{0,0,180}{Answer:}\\
\rule{\linewidth}{0.5pt}
\begin{center}
    \textbf{(b) An example of zero-shot prompting for XfundQA dataset}
\end{center}
\vspace{-5pt}
The following is a form composed of key-value pairs: "\{context\}". Please answer according to the given form. \\
Note: The value usually appears near the key. Think carefully and answer with a few words. \\
\textcolor[RGB]{180,0,0}{Question:} What is the value of the key "\{question\}"? \\
\textcolor[RGB]{0,0,180}{Answer:}\\
\rule{\linewidth}{0.5pt}
\begin{center}
    \textbf{(c) An example of zero-shot prompting for DocVQA dataset}
\end{center}
\vspace{-5pt}
Given the context: \\
\{context\} \\
\textcolor[RGB]{180,0,0}{Use few words to answer the question:} \{question\} \\
\textcolor[RGB]{0,0,180}{Answer:}\\
\rule{\linewidth}{0.5pt}
\begin{center}
    \textbf{(d) An example of zero-shot prompting for FeTaQA dataset}
\end{center}
\vspace{-5pt}
Given a table: \\
\{context\} \\
Answer questions about the table. \\
Note: think step by step. \\
\textcolor[RGB]{180,0,0}{Question:} \{question\} \\
\textcolor[RGB]{0,0,180}{Answer:}\\
\rule{\linewidth}{0.5pt}
\begin{center}
    \textbf{(e) An example of 3-shot prompting for rephrasing answers in DocVQA dataset}
\end{center}
\vspace{-5pt}
Given the question and answer pair, rephrase the answer to provide the most straightforward response to the question with few words in English. \\
\\
Example 1:  \\
\textcolor[RGB]{180,0,0}{Question:} What is the name of the person in the CC field? \\
\textcolor[RGB]{0,0,180}{Answer:} The name of the person in the CC field is Jo Spach. \\
\textcolor[RGB]{0,180,0}{Rephrased answer:} Jo Spach \\
\\
Example 2:  \\
\textcolor[RGB]{180,0,0}{Question:} What is the given document about? \\
\textcolor[RGB]{0,0,180}{Answer:} The given document appears to be a summary of an evaluation survey conducted by Telmark in a particular monthly region in 2014. The survey aimed to evaluate the effectiveness of Telmark's promotional programs in the region. The document provides information on various aspects of the survey, including the number of stores that received promotional materials, the percentage of stores that placed the materials in a visible location, and the number of stores that participated in the promotion. Additionally, the document includes information on the wholesale accounts sold by Telmark in the region and the percentage of accounts that refused the promotion. \\
\textcolor[RGB]{0,180,0}{Rephrased answer:} region monthly telmark program evaluation survey \\
\\
Example 3:  \\
\textcolor[RGB]{180,0,0}{Question:} What is the \% of Employees in 2012 based on graph 'Distribution of Value-Added'? \\
\textcolor[RGB]{0,0,180}{Answer:} Based on the graph 'Distribution of Value-Added', it can be observed that the percentage of employees in 2012 is around 80\%. \\
\textcolor[RGB]{0,180,0}{Rephrased answer:} 80\% \\
\\
Now rephrase the answer based on the QA pair: \\
\textcolor[RGB]{180,0,0}{Question:} \{question\} \\
\textcolor[RGB]{0,0,180}{Answer:} \{answer\} \\
\textcolor[RGB]{0,180,0}{Rephrased answer:}\\
\rule{\linewidth}{0.5pt}
\begin{center}
    \textbf{(f) Prompt template for Llama2}
\end{center}
\vspace{-5pt}
\begin{lstlisting}[basicstyle=\footnotesize\ttfamily]
<s>[INST] <<SYS>>
{system prompt}
<</SYS>>
{instruction} [/INST]
\end{lstlisting}
\rule{\linewidth}{1pt}
\vspace{1pt}
\captionof{figure}{Prompt designs for different datasets.}
\vspace{10pt}
\label{fig:8}

\section{Examples of different table encoding methods} \label{table-encoding}
Figure \ref{fig:5} shows examples of different table encoding methods. The widely used table encoding methods include: arranging data in array format (Array), using unique identifiers to distinguish between headers and rows (Linear), and formatting each element as a column-row-value triplet to form a list (Triple).

\clearpage
\noindent\begin{minipage}{\textwidth}
\centering
\rule{15cm}{1pt}
\begin{center}
    \textbf{(a) Array}
\end{center}
\vspace{-5pt}
\begin{lstlisting}[basicstyle=\footnotesize\ttfamily]
                [['Year', 'Title', 'Role', 'Channel'],
                ['2015', 'Kuch Toh Hai Tere Mere Darmiyaan', 'Sanjana Kapoor', 'Star Plus'],
                ['2016', 'Kuch Rang Pyar Ke Aise Bhi', 'Khushi', 'Sony TV'],
                ['2016', 'Gangaa', 'Aashi Jhaa', '\&TV']]
\end{lstlisting}

\rule{15cm}{0.5pt}
\begin{center}
    \textbf{(b) Linear}
\end{center}
\vspace{-5pt}
\begin{lstlisting}[basicstyle=\footnotesize\ttfamily]
                [HEAD] Year | Title | Role | Channel
                [ROW] 1 2015 | Kuch Toh Hai Tere Mere Darmiyaan | Sanjana Kapoor | Star Plus
                [ROW] 2 2016 | Kuch Rang Pyar Ke Aise Bhi | Khushi | Sony TV
                [ROW] 3 2016 | Gangaa | Aashi Jhaa | \&TV
\end{lstlisting}

\rule{15cm}{0.5pt}
\begin{center}
    \textbf{(c) Triplet}
\end{center}
\vspace{-5pt}
\begin{lstlisting}[basicstyle=\footnotesize\ttfamily]
                Row1 | Year | 2015
                Row1 | Title | Kuch Toh Hai Tere Mere Darmiyaan
                Row1 | Role | Sanjana Kapoor
                Row1 | Channel | Star Plus
                Row2 | Year | 2016
                Row2 | Title | Kuch Rang Pyar Ke Aise Bhi
                Row2 | Role | Khushi
                Row2 | Channel | Sony TV
                Row3 | Year | 2016
                Row3 | Title | Gangaa
                Row3 | Role | Aashi Jhaa
                Row3 | Channel | \&TV
\end{lstlisting}

\rule{15cm}{0.5pt}
\begin{center}
    \textbf{(d) Ours}
\end{center}
\vspace{-5pt}
\begin{lstlisting}[basicstyle=\footnotesize\ttfamily]
                Year     Title                             Role                      Channel
                2015     Kuch Toh Hai Tere Mere Darmiyaan  Sanjana Kapoor            Star Plus
                2016     Kuch Rang Pyar Ke Aise Bhi        Khushi                    Sony TV
                2016     Gangaa                            Aashi Jhaa                \&TV
\end{lstlisting}

\rule{15cm}{1pt}
\vspace{10pt}
\captionof{figure}{Different table encoding methods: "Array," which transforms the original array table data into string format; "Linear," which employs distinct identifiers to differentiate headers and rows; "Triplet," which formats each element as a col-row-value triplet to create a list; and "Ours," which utilizes spaces and line breaks to align and separate elements within the table.}
\vspace{-130pt}
\label{fig:5}
\end{minipage}
\end{appendices}

\end{document}